\documentclass[runningheads]{llncs}
\usepackage{graphicx}
\usepackage{graphbox}
\newcommand{\p}[1]{p. #1} 
\newcommand{\pp}[1]{pp. #1} 
\begin{document}
\title{Introducing the diagrammatic semiotic mode}
\author{Tuomo Hiippala\inst{1}\orcidID{0000-0002-8504-9422} \and John A. Bateman\inst{2}\orcidID{ 0000-0002-7209-9295}
}
\authorrunning{Hiippala and Bateman}
\institute{University of Helsinki, Helsinki, Finland \and Bremen University, Bremen, Germany  \\
\email{tuomo.hiippala@helsinki.fi} \\
\email{bateman@uni-bremen.de}
}
\maketitle              % typeset the header of the contribution
\begin{abstract}
As the use and diversity of diagrams across many disciplines grows, there is an increasing interest in the diagrams research community concerning how such diversity might be documented and explained. In this article, we argue that one way of achieving increased reliability, coverage, and utility for a general classification of diagrams is to draw on recently developed semiotic principles developed within the field of multimodality. To this end, we sketch out the internal details of what may tentatively be termed the \emph{diagrammatic semiotic mode}. This provides a natural account of how diagrammatic representations may integrate natural language, various forms of graphics, diagrammatic elements such as arrows, lines and other expressive resources into coherent organisations, while still respecting the crucial diagrammatic contributions of visual organisation. We illustrate the proposed approach using two recent diagram corpora and show how a multimodal approach supports the empirical analysis of diagrammatic representations, especially in identifying diagrammatic constituents and describing their interrelations in a manner that may be generalised across diagram types and be used to characterise distinct kinds of functionality.

\keywords{Diagrams \and Semiotics \and Multimodality \and Corpora \and Annotation.}
\end{abstract}

\section{Introduction}

Diagrams appear to be playing an ever greater range of roles in a similarly increasing range of application contexts. Several authors have consequently called for more efforts to characterise this diversity so as to establish more finely articulated accounts of just what kinds of diagrams there are and how they might serve different communicative and cognitive functions. Norman, for example, considers a two-dimensional characterisation in terms of `discretion' and `assimilability' to distinguish more clearly the role of diagrams among the traditionally drawn categories of descriptions, diagrams, and depiction \cite{Norman2000-diagram-types}. Smessaert and Demey offer a typology of types of diagrams used in linguistics, focusing on more content-oriented `linguistic parameters' and specific semiotically-motivated diagrammatic parameters distinguishing iconic and symbolic representations \cite{SmessaertDemey2018}. Johansen \emph{et al.} offer a typology of mathematical diagrams based on their use by mathematicians, identifying `resemblance', `abstract' and  `Cartesian' diagrams \cite{Johansen-etal2018-diagram-typology}. And Purchase and colleagues  propose a multidimensional classification of infographics by analysing how users grouped a selection of 60 infographics \cite{Purchase-etal2018-infographics}.

Work of this kind raises important research questions, including questions of the particular cognitive capabilities demanded or supported by distinct diagram types \cite[\p{107}]{Johansen-etal2018-diagram-typology}, the ways in which diagram usage has developed and expanded over time \cite[\p{106}]{Johansen-etal2018-diagram-typology}, and how studies might be extended to address entire collections of diagrams, thereby focusing and organising lines of research on a broader scale than hitherto  \cite[\p{210}] {Purchase-etal2018-infographics}. Strengthening the empirical basis for diagrams research by drawing on broader sets of examples and providing more finely articulated characterisations of the properties of diagrams that go beyond existing categories, such as the fundamental distinctions offered by Peirce in terms of iconicity, indexicality and symbolicity and so on, are consequently now well established as aims. As Johansen \emph{et al.} argue, Peirce's functional definition, particular of iconicity, ``is too broad and does not allow for making cognitively and practically meaningful distinctions in the category of diagrams" \cite[\p{107}]{Johansen-etal2018-diagram-typology}. Nevertheless, developing convincing classifications of diagrams -- even in specific areas -- has proved challenging. 

Both the conceptually-based development of frameworks \cite{Norman2000-diagram-types,tversky2017,SmessaertDemey2018,Johansen-etal2018-diagram-typology,EngelhardtRichards2018} and more bottom-up clustering based on human judgements \cite{Purchase-etal2018-infographics} continue to face issues of exhaustivity, discriminability, and reliability. Several significant problems are noted by Johansen \emph{et al.}:
\begin{quote}
  ``Although such a classification gave a more fine-grained resolution in the diagram classification \ldots it turned out to be difficult to carry out in practice \ldots Consequently, counting the number of diagrams of a specific type requires making judgements based on the visual appearance of diagrams to classify them correctly." \cite[p. 116]{Johansen-etal2018-diagram-typology}  
\end{quote}
Indeed, ``we also encountered cases where we had to involve the textual or intellectual context of diagrams to classify them, and in other cases, we could only give educated guesses.'' \cite[\p{116}]{Johansen-etal2018-diagram-typology}. Moreover, even in Purchase \emph{et al.}'s `user-based' classification, the Likert scores used to evaluate the infographics were found to be poor predictors of class with ``their values bear[ing] little relation to the groupings created by the participants'' \cite[\p{216}]{Purchase-etal2018-infographics}; again, visual grouping of diagrams appeared to give stronger results.

Consequently, despite the generally positive reports and indications of considerable utility of working with diagram collections, it is less clear whether the classifications proposed to date can be scaled-up in a reliable fashion. Certain gaps appear to occur due to the context-dependent nature of any functional categories employed. In this article, therefore, we address these issues from a complementary perspective and argue that the resulting multiply-dimensioned classification promises a more robust approach to diagram classification, capable not only of respecting visual appearance in a systematic and philosophically sound fashion, but also of providing a principled approach to context-dependent interpretations as well. 

This approach draws on classification work in the field of multimodality research, an emerging discipline that studies how communication builds on appropriate combinations of multiple modes of expression, such as natural language, illustrations, drawings, photography, gestures, layout and many more \cite{batemanetal2017}. One product of this work is a battery of theoretical concepts that strongly support \emph{empirical} analysis of complex communicative situations and artefacts. We describe how this can now be applied directly to the analysis of diagrams within the context of the challenges set out above. For this, we define what we tentatively term the \emph{diagrammatic semiotic mode}. This bridges discussions in the diagrams research and multimodality communities by introducing an explicitly multimodal, discourse-oriented perspective to diagrams research. We illustrate this in relation to two recently published multimodal diagram corpora. 

\section{A multimodal perspective on diagrams}

\begin{figure}[h!]
\includegraphics[width=1\textwidth]{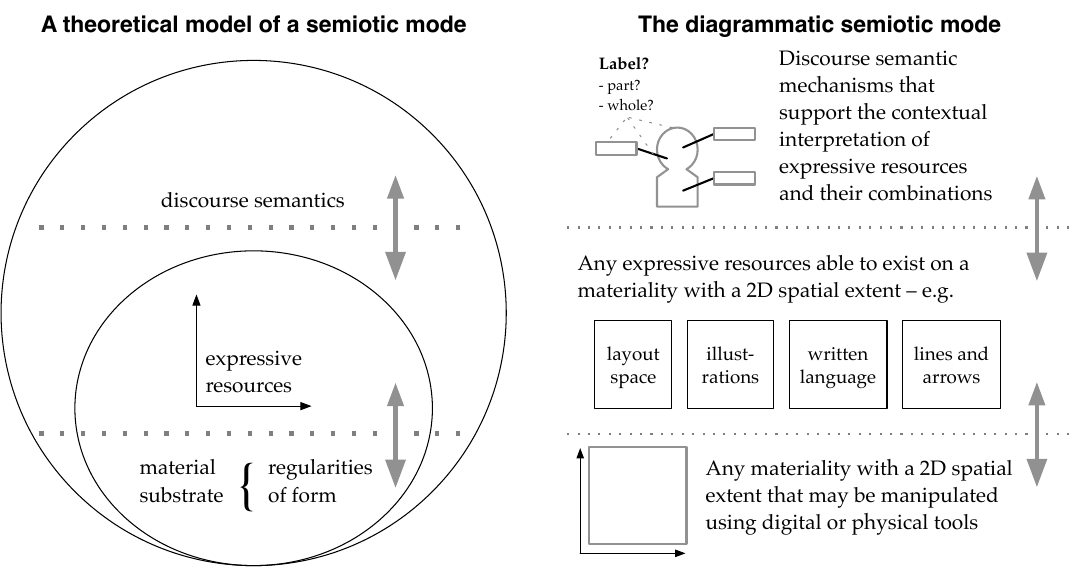}
\caption{A theoretical model of a semiotic mode and a sketch of the fundamentals for a diagrammatic semiotic mode \cite[p. 4]{hiippalabateman2021}}
\label{fig:semiotic_mode}
\end{figure}

The framework of multimodality adopted here offers a common set of concepts and an explicit methodology for supporting empirical research regardless of the `modes' and materials involved \cite{batemanetal2017}. The result is capable of addressing all forms of multimodal representation, including diagrammatic representations of all kinds. The core theoretical concept within the framework is that of the \emph{semiotic mode},  a graphical definition of which we show on the  left-hand side of Figure \ref{fig:semiotic_mode}. Here we see three distinct `semiotic strata' that the model claims are always needed for a fully developed semiotic mode to operate \cite{Bateman2011-modes}.

Starting from the lower portion of the inner circle, the model requires that all semiotic modes work with respect to a specified \emph{materiality} which a community of users regularly `manipulates' in order to leave traces for communicative purposes; second, these traces are organised (paradigmatically and syntagmatically) to form \emph{expressive resources} that characterise the material distinctions that are specifically pertinent for the semiotic mode at issue; and finally, those expressive resources are mobilised in the service of communication by a corresponding \emph{discourse semantics}, whose operation we show in a moment. 

In general, no ordering is imposed on the flow of information across these three strata, although methodologically it can often be beneficial to begin with the more observable material traces. Different semiotic modes also provide differing degrees of constraint at the various levels: for example, whereas the semiotic mode of verbal language offers substantial form-driven constraints guiding discourse interpretation, pictorial semiotic modes often require more discourse constraints when selecting between perceptually plausible readings -- Bateman, Wildfeuer and Hiippala \cite[\p{33}]{batemanetal2017} discuss an example offered by Gombrich \cite[\p{7}]{Gombrich60} further from this perspective, showing how variability in interpretation is naturally supported. Finally, the model places no restrictions on the kinds of materiality that may be employed; for current purposes, however, we illustrate the approach by focusing on static two-dimensional diagrams. As Bateman \cite{bateman2021} shows, however, the approach generalises equally to both dynamic and 3D cases.

Building on this scheme,  we set out on the right-hand side of the figure an initial characterisation of the specific properties of the diagrammatic semiotic mode. The 2D materiality of this mode not only allows the creation of spatial organisations in the form of layout, but is also a prerequisite for realising many of the further expressive resources commonly mobilised in diagrams, such as written language and arrows, lines, glyphs and other diagrammatic elements, which also inherently require (at least) a 2D material substrate. An example of the corresponding expressive resources typical of the diagrammatic mode is offered by the ``meaningful graphic forms'' identified by Tversky \textit{et al.} \cite[\p{222}]{tverskyetal2000}, such as circles, blobs and lines. These can also be readily combined into larger syntagmatic organisations in diagrams such as route maps, as Tversky et al. illustrate \cite[\p{223}]{tverskyetal2000}; Engelhardt and Richards offer a similar set of `building blocks' \cite[\p{201}]{EngelhardtRichards2018}. However, theoretically, the diagrammatic semiotic mode can in fact draw on any expressive resource capable of being realised on a materiality with a 2D spatial extent, although in practice these choices are constrained by what the diagram attempts to communicate and the sociohistorical development of specific multimodal \emph{genres} by particular communities of practice \cite{bateman2008,hiippala2015a}. Finally, it is the task of the third semiotic stratum of discourse semantics to make the use of expressive resources interpretable in context. 

Embedding expressive resources into the discourse organisations captured by a discourse semantics is crucial to our treatment and a key extension beyond traditional semiotic accounts. Essentially, this enables the account to do full justice to the Peircean embedding of iconic forms within conventionalised usages \cite{Stjernfelt2007-diagrammatology}. It is this addition that explains formally how (and why) fundamental graphic forms, such as those identified by Engelhardt and Richards, Tversky \textit{et al}. and others, may receive different interpretations in different contexts of use -- a problem noted for several of the classifications introduced above -- while also allowing certain intrinsic properties of those forms (such as connectivity and directionality) to play central roles in finding interpretations as well. The combination of materiality, expressive forms and discourse interpretations then provides a robust foundation for considerations of diagrammatic reasoning quite generally. 

\section{Multimodal diagram corpora}

We now illustrate the potential of a characterisation of diagrams drawing on our multimodal framework for dealing with collections of diagrams by considering two concrete, interrelated diagram corpora: AI2D \cite{kembhavietal2016} and AI2D-RST \cite{hiippalaetal2019-ai2d}. These corpora build on one another, as AI2D-RST covers a subset of AI2D. We describe the corpora and how they have been characterised and show how an increasing orientation to multimodality successively raises the accuracy and utility of the classification applied. We will argue that the characterisation provided supports a general methodology for building classifications for collections of diagrams. 

\subsection{The Allen Institute for Artificial Intelligence Diagrams dataset} \label{sec:ai2d}

The Allen Institute for Artificial Intelligence Diagrams dataset (AI2D) was developed to support research on computational processing of diagrams \cite{kembhavietal2016}. AI2D contains a total of 4903 diagrams that represent topics in elementary school natural sciences, ranging from life and carbon cycles to human physiology and food webs, to name just a few of the 17 categories in the dataset. Because the diagram images were scraped from the web using school textbook chapter headings as search terms, the corpus covers a wide range of diagrams created by producers with various degrees of expertise with the diagrammatic semiotic mode, such as students, teachers and professional graphic designers. As the diagrams have been removed from their original context during scraping, little may be said about the medium they originated in. For this reason, it may be suggested that AI2D \emph{approximates} how diagrams are used in learning materials realised using various media.

AI2D models four types of diagram elements: text, blobs (graphic elements), arrows and arrowheads. Although these elements cover the main expressive resources mobilised in these diagrams, no further distinctions are made between drawings, illustrations, photographs and other visual expressive resources \cite{hiippalabateman2021}. Each diagram in the dataset is nevertheless provided with several layers of description. Instances of the four diagram element types were first segmented from the original diagram images by crowdsourced workers \cite[\p{243}]{kembhavietal2016}. The elements identified during this \emph{layout segmentation} provide a foundation for a \emph{Diagram Parse Graph} (DPG), which represents the diagram elements as nodes and semantic relations between elements as edges. Ten relations are used, drawing from the framework proposed by Engelhardt \cite{engelhardt2002}. Crowdsourcing annotations is a promising way of creating larger-scale collections of classified diagrams -- it also, however, demands that the classifications constructed are sufficiently clear and well-defined to avoid the drawbacks observed by other approaches reported in the introduction above \cite[\pp{683--684}]{hiippalaetal2019-ai2d}.

\begin{figure}[p]
\centering
\includegraphics[width=0.68\textwidth]{4210.png}
\includegraphics[width=0.68\textwidth]{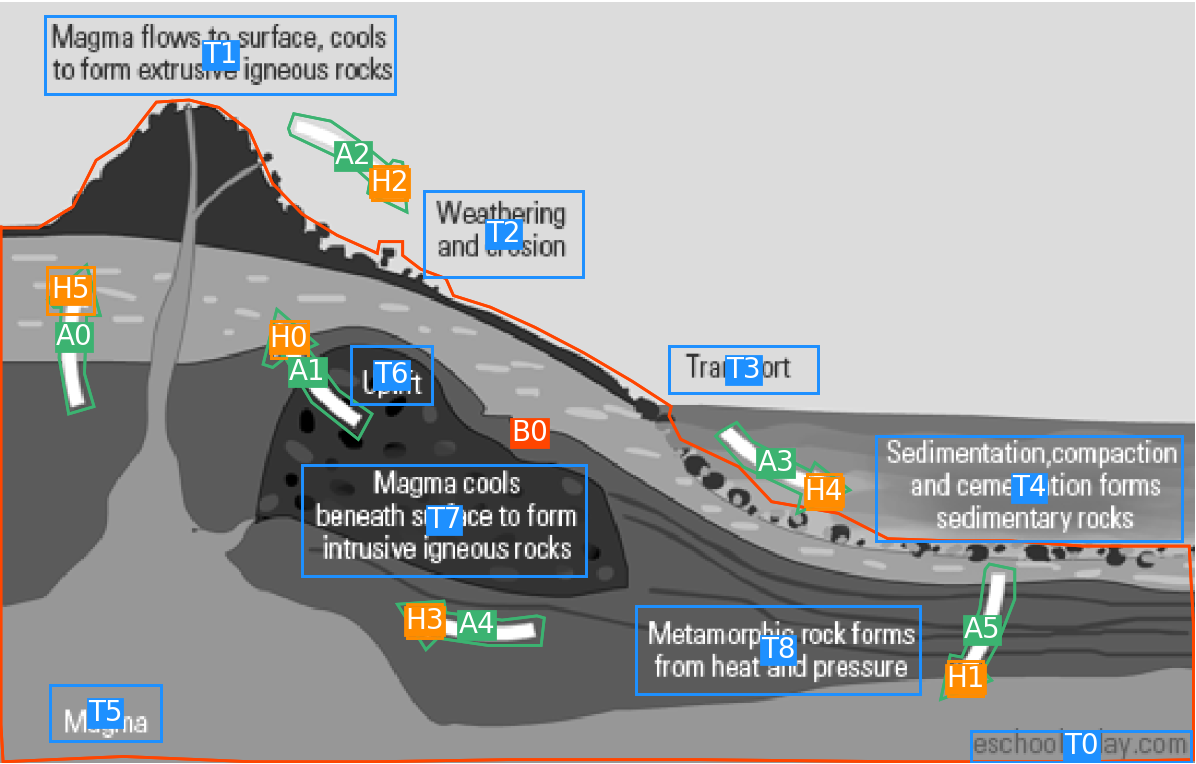} \\[0.5cm]
\includegraphics[width=0.68\textwidth]{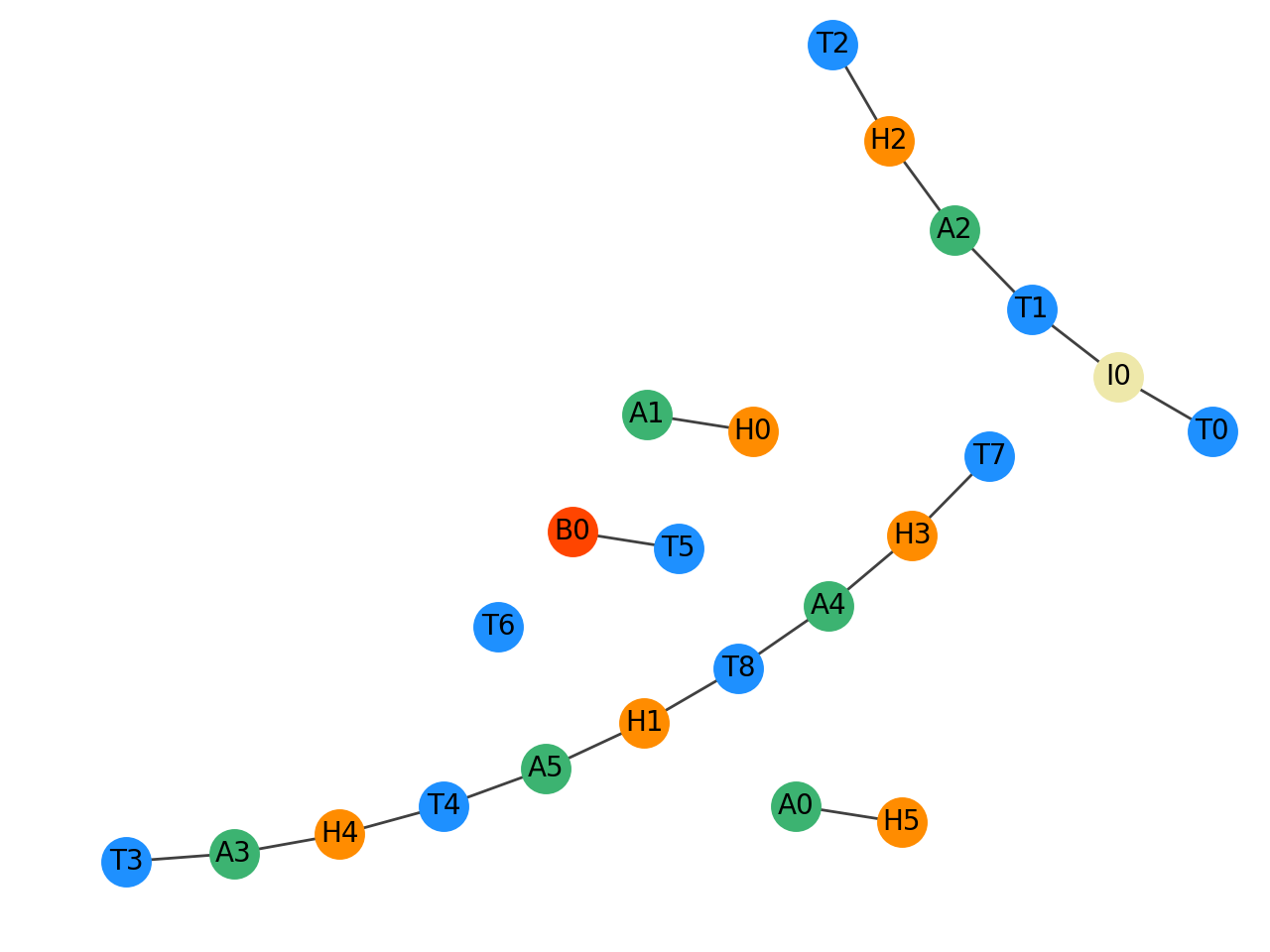} 
\caption{Original diagram image (top), layout segmentation (middle) and Diagram Parse Graph (bottom) for diagram \#4210 in the AI2D corpus. In the layout segmentation, the original image has been converted into grayscale to highlight the crowd-sourced layout segmentation. Each layout segment is coloured according to diagram element type (blue: text; red: blob; arrow: green; arrowhead: orange) and assigned a unique identifier. These colours and identifiers are carried over to the Diagram Parse Graph.}
\label{fig:ai2d_example}
\end{figure}

Figure \ref{fig:ai2d_example} shows as an example the treatment given to a diagram originally scraped from the web, diagram \#4210 in AI2D. Below the original shown at the top of the figure, we see the diagram's crowdsourced layout segmentation and, at the bottom, its corresponding DPG. The original diagram represents a rock cycle, that is, transitions between different types of rock, using a combination of an illustration (a cross-section) whose parts are described using written language. These parts set up the stages of the rock cycle, which are then related to one another using arrows.

For the formation of the AI2D corpus, annotators were instructed to identify units and relationships. As the resulting layout segmentation image in the middle of the figure shows, text blocks and arrowheads were segmented using rectangular bounding boxes, whereas more complex shapes for arrows and various types of graphics were segmented using polygons. The layout segmentation illustrates a common problem with crowdsourced annotations: annotators tend to segment diagrams to quite uneven degrees of detail. Here the entire cross-section is assigned to a single blob (B0), although a more accurate description would be to segment separate parts of the cross-section, such as magma and various layers of rock. We will see shortly how such omissions readily compromise the accurate description of semantic relations in the DPG.

Referring again to the layout segmentation and DPG in the figure, we can see for example that the semantic relations carried by the edges in the DPG cover \textsc{arrowHeadTail} between arrow A2 and arrowhead H2 in the upper part of the diagram, which together act as a connector in an \textsc{interObjectLinkage} relation between text blocks T1 (`Magma flows to surface ...') and T2 (`Weathering and erosion'). As these relations illustrate, Engelhardt's \cite{engelhardt2002} relations cover local relations holding between diagram elements that are positioned close to one another or connected using arrows or lines. They neglect, however, the relations needed to describe the global organisation of the diagram, that is, relations between units that are made up of multiple elements \cite[\p{239}]{kembhavietal2016}. Crowdsourcing coherent graph-based descriptions of diagrams thus turns out to be a challenging task. AI2D DPGs often include isolated nodes and multiple connected components, as exemplified by the DPG in Figure \ref{fig:ai2d_example}. Furthermore, the DPG does not feature a cyclic structure, although the diagram clearly describes a rock \emph{cycle}. The AI2D annotation scheme provides the relation definitions necessary for describing this process in principle, such as \textsc{interObjectLinkage} and \textsc{intraObjectRegionLabel} \cite[\p{239}]{kembhavietal2016}, but annotators following a more shallow visual grouping would not be led to this option from the diagram at hand. As we shall see, this is one of many problems that an explicit discourse semantic orientation can address. 

The crowdsourced annotators were not explicitly instructed to decompose cross-sections or other visual expressive resources capable of demarcating meaningful regions, which results in an insufficiently detailed layout segmentation. The blob B0, which covers the entire cross-section shown in the diagram, is as a consequence not segmented into its component parts -- i.e., the stages of the rock cycle with labels such as `Magma' (T5) and `Metamorphic rock forms from heat and pressure' (T8) even though each of these picks out a  particular region of the cross-section through visual \emph{containment} \cite[\p{47}]{engelhardt2002} as necessary for defining the stages of the cycle. The cross-section (B0) instead constitutes a single unit and so an otherwise applicable relation such as \textsc{intraObjectRegionLabel} cannot be used to pick out corresponding regions simply because those regions are not present in the inventory of identified elements. As such, the description is not sufficiently detailed to represent a cyclic structure.

These challenges relating to decomposing diagrammatic representations relate to the well-known problem of identifying `units' mentioned above and discussed in multimodality theory for many visually-based semiotic modes. In general an annotator, be that an expert analyst or a crowdsourced non-expert worker, will not know on purely visual grounds whether it is necessary, or beneficial, to segment areas presented in a diagram. As we shall see, this is precisely where we need to engage a corresponding notion of discourse semantics for the semiotic mode at issue. The discourse semantics  simultaneously supports \emph{decomposing} larger units into component parts and \emph{resolving} their potential interrelations, always with the goal of maximising \emph{discourse coherence} \cite[\p{377}]{batemanwildfeuer2014b}. In the next section, we show how this approach can be used for a more effective design of a multimodal corpus of diagrams.

\subsection{AI2D-RST -- a multimodally-motivated annotation schema}

The second corpus considered here, AI2D-RST, covers 1000 diagrams taken from the AI2D corpus and is annotated using a new schema by experts specifically trained in the use of that schema \cite{hiippalaetal2019-ai2d}. The primary goal here was precisely to compare the original corpus, with its style of classification, to a corpus adopting a classification more explicitly anchored into the requirements raised by the diagrammatic semiotic mode. The development of AI2D-RST was motivated by the observation that the AI2D annotation schema introduced above conflates descriptions of different types of multimodal structure, such as implicit semantic relations and explicit connections signalled using arrows and lines, into a single DPG \cite{hiippalaorekhova2018}. These can now be separated multimodally so as to better understand how such structures contribute to diagrammatic representations.

\begin{figure}[h!]
\includegraphics[width=0.5\textwidth]{segmentation_4210.png} 
\includegraphics[width=0.5\textwidth]{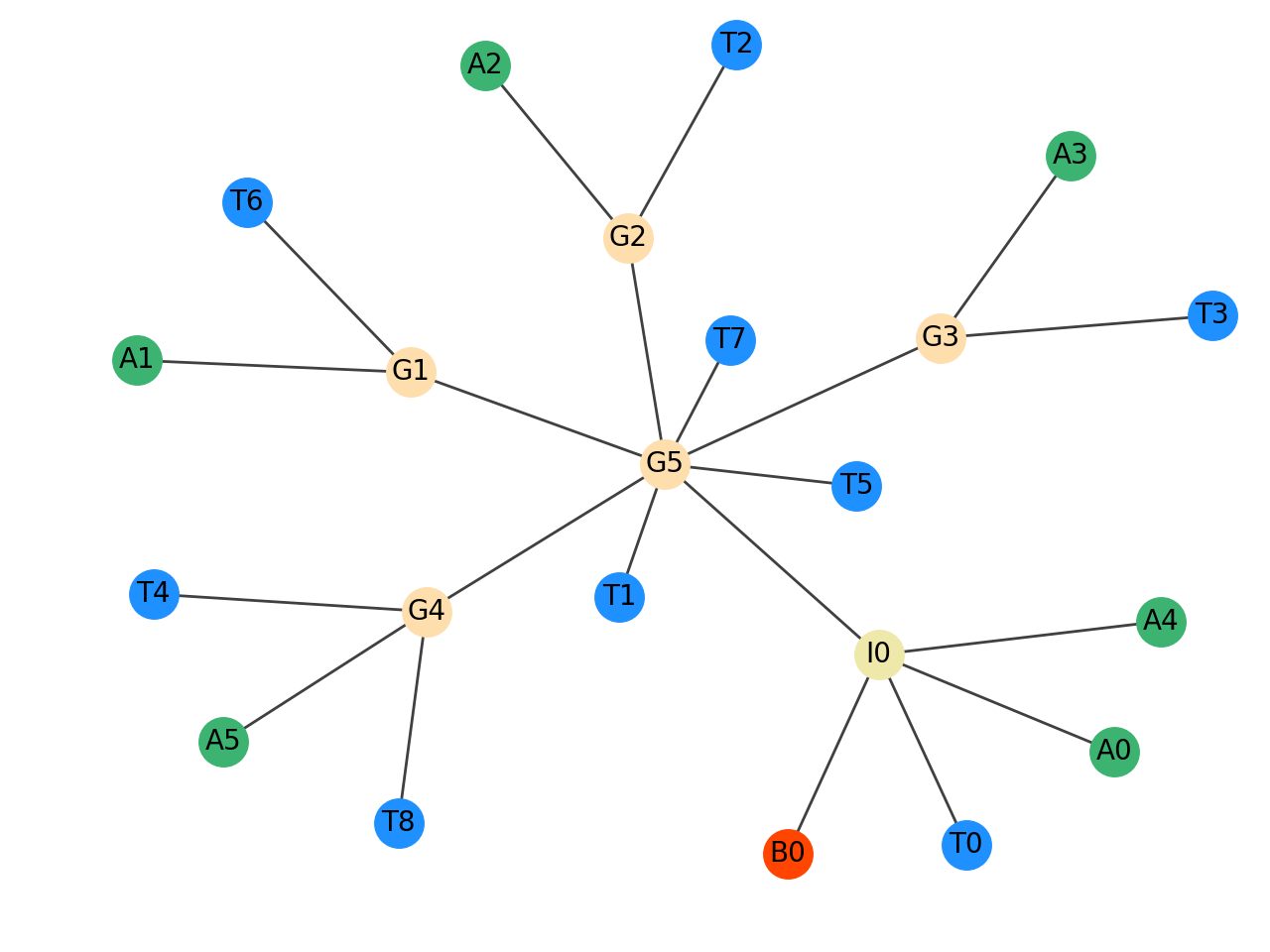} 
\includegraphics[width=0.5\textwidth]{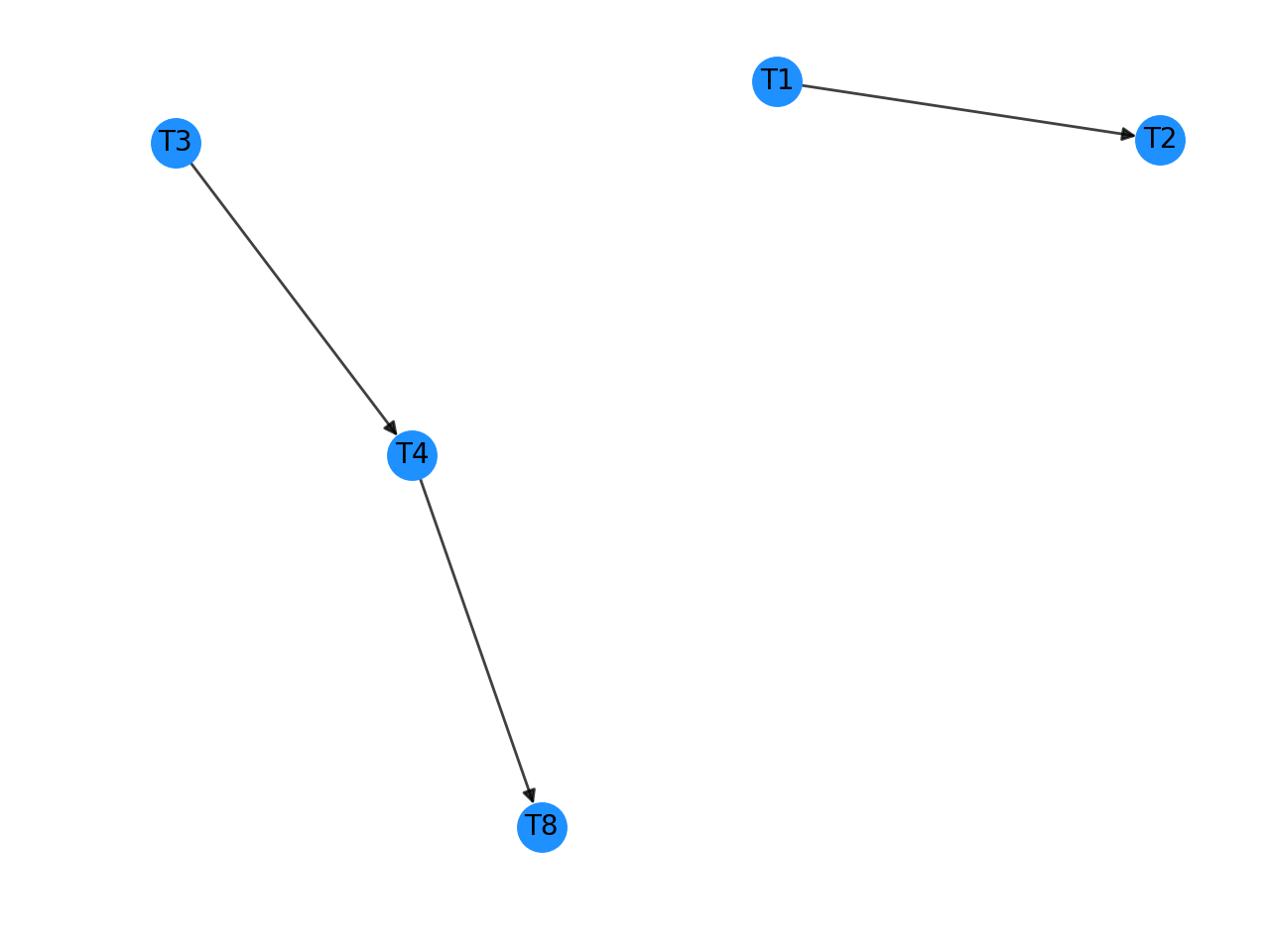} 
\includegraphics[width=0.5\textwidth]{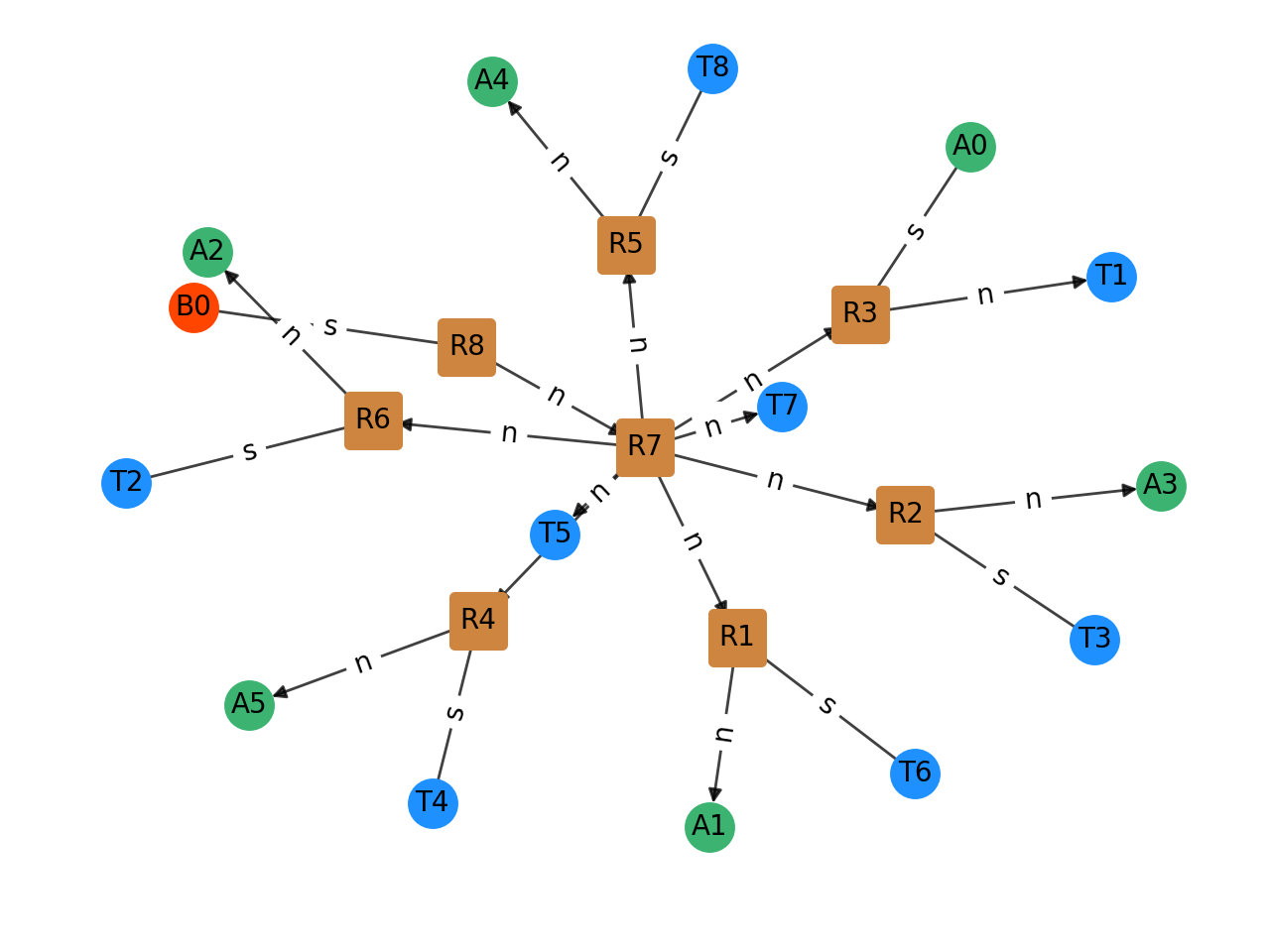} 
\caption{The original crowd-sourced layout segmentation from AI2D (top left) and AI2D-RST grouping (top right), connectivity (bottom left; with two subgraphs) and discourse structure (bottom right) graphs for diagram \#4210. Note that unlike AI2D, AI2D-RST does not model arrowheads as individual units, which is why they are absent from the graphs. This information can be retrieved from the original AI2D annotation if needed.}
\label{fig:ai2d-rst_example}
\end{figure}

To achieve this, AI2D-RST  represents each diagram using three distinct graphs corresponding to three distinct, but mutually complementary, layers of multimodally motivated annotations: grouping, connectivity and discourse structure. Figure \ref{fig:ai2d-rst_example} shows examples of all three graphs for the diagram from Figure \ref{fig:ai2d_example}. To begin, the grouping layer (top right) organises diagram elements that are likely to be perceived as belonging together into visual perceptual groups loosely based on Gestalt properties \cite{ware2012}. The resulting organisation is represented using a hierarchical tree graph, with grouping nodes with the prefix `G' added to the graph as parents to nodes grouped together during annotation. Such grouping nodes can be picked up in subsequent annotation layers to refer to a group of diagram elements and thereby serve as a foundation for the description of both the connectivity and discourse structure layers.

The connectivity layer (bottom left) is represented using a cyclic graph, in which edges represent visually explicit connections signalled using arrows and lines in the diagram. As the connectivity graph in Figure \ref{fig:ai2d-rst_example} shows, it is important that these cover explicit connections only since this reveals the diagram to leave several gaps in its characterisation of the rock cycle, namely between the stages represented using text blocks T7 (`Magma cools beneath surface ...') and T1 (`Magma flows to surface ...'), and between T2 (`Weathering and erosion) and T3 (`Transport'). It is consequently left to the viewer to fill in such connections during discourse interpretation. Not including such connections  in the description of connectivity allows us to capture discrepancies between explicit visual signals, such as arrows and lines, and implicit meanings that are only derivable from the discourse structure.

In AI2D-RST, such implicit discourse relations are handled by the third layer, that of discourse structure, which uses Rhetorical Structure Theory (RST) \cite{mannthompson1988,taboadamann2006a} to describe semantic relations between diagram elements. RST was originally developed as a theory of text organisation and coherence in the 1980s \cite{mannthompson1988} and has frequently been applied subsequently to the description of discourse semantics in multimodality research as well \cite{bateman2008}. Originally, RST attempted to describe why well-formed texts appear coherent, or why individual parts of a text appear to contribute towards a common communicative goal \cite{taboadamann2006a}, and so this is a relatively natural perspective to take on diagrams and other forms of multimodal communication. RST defines a set of `rhetorical relations' that are intended to capture the communicative intentions of the designer, as judged by an analyst. AI2D-RST applies these relations to diagrams from the AI2D dataset to provide an alternative annotation schema offering a more multimodally informed description of the intended functions of diagrammatic representations \cite{hiippalaetal2019-ai2d}. 

The relations defined by RST are added to the discourse structure graph of diagrams in the corpus as nodes prefixed with the letter `R' as shown in the graph bottom right in Figure \ref{fig:ai2d-rst_example};  the edges of the graph describe which role an element takes in the discourse relation, namely nucleus (`n') or satellite (`s'). This notion of \emph{nuclearity} is a key criterion in definitions of semantic relations in RST. Following the original RST definitions, AI2D-RST represents the discourse structure layer using a strict tree graph: if a diagram element is picked up as a part of multiple rhetorical relations, a duplicate node is added to the graph to preserve the formal requirement of tree structure.

In Figure \ref{fig:ai2d-rst_example}, the specific rhetorical relations in the bottom right graph include \textsc{identification} (R1--R6), \textsc{cyclic sequence} (R7) and \textsc{background} (R8). Since AI2D-RST still builds on the inventory of diagram elements provided by the original layout segmentation in AI2D, this requires some compromises in the RST analysis. Here the original annotator of the diagram had concluded that most text instances serve to identify what the arrows stand for, namely stages of the rock cycle. The image showing the cross-section (B0), in turn, is placed in a \textsc{background} relation  to the \textsc{cyclic sequence} relation. The definition of a \textsc{background} relation \cite{mannthompson1988} states that the satellite (B0) increases the ability to understand the nucleus (R7), which is the top-level relation assigned to the diagram's representation of the entire cycle.

Although this offers a first  incremental step for including discourse information in a diagram corpus, building directly on the original AI2D corpus and its segmentation is also a severe limitation. In fact, this offers only a rather crude description of the discourse structure of the diagram in Figure \ref{fig:ai2d-rst_example} because the cross-section B0 is actually providing far more information. This information is crucial for understanding what the diagram is attempting to communicate \emph{but we cannot know that such a decomposition is necessary without considering the rhetorical discourse organisation of the diagram as a whole}. The particular decomposition of diagrams must often be pursued in a top-down direction therefore, emphasising the discourse structure from the outset \cite{batemanwildfeuer2014b}. Without methodologically prioritising the analysis of discourse structure, it is difficult to know which aspects of the diagrammatic mode are being drawn on and which elements should actually be included in the description of discourse structure. 

This is one of the basic problems underlying several of the limitations discussed for previous diagram classifications above. A visually-accessible cross-section such as the one shown in Figure \ref{fig:ai2d_example} is, in fact, very likely to use \emph{illustration} or other expressive resources capable of representing and demarcating meaningful regions in 2D layout space \cite{richards2017}. This possibility makes the question of whether the capability is actually being drawn on pertinent and, if the capability is used, raises further the issue of the extent to which the illustration  must be decomposed so as to achieve the inventory of elements needed for making appropriate inferences about the discourse structure. Analytical problems arising from the original layout segmentation are consequently still being propagated from AI2D to AI2D-RST.

\subsection{Next step: adding discourse-driven decomposition to AI2D-RST}

\begin{figure}[h!]
\includegraphics[width=0.5\textwidth]{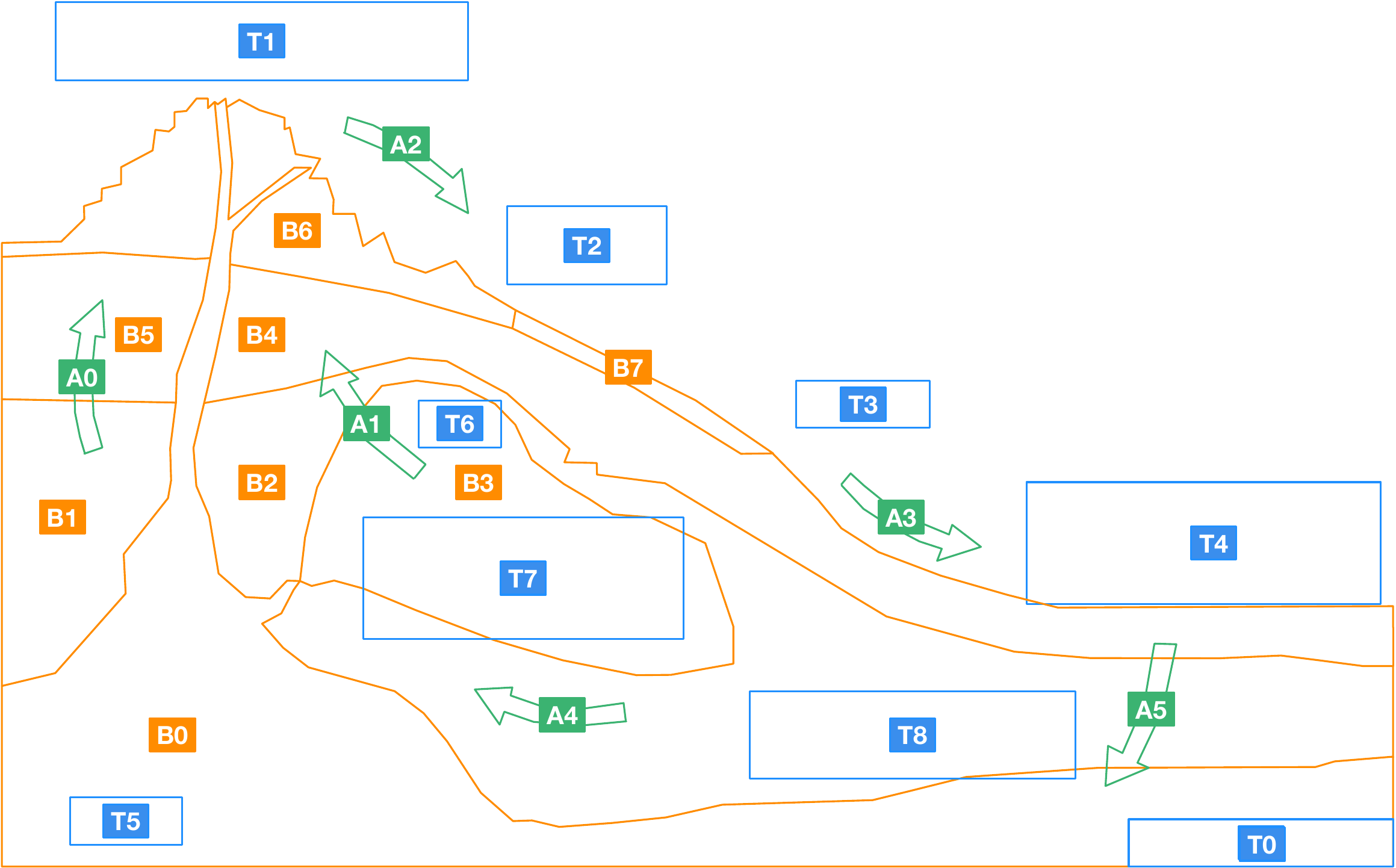} 
\includegraphics[width=0.5\textwidth]{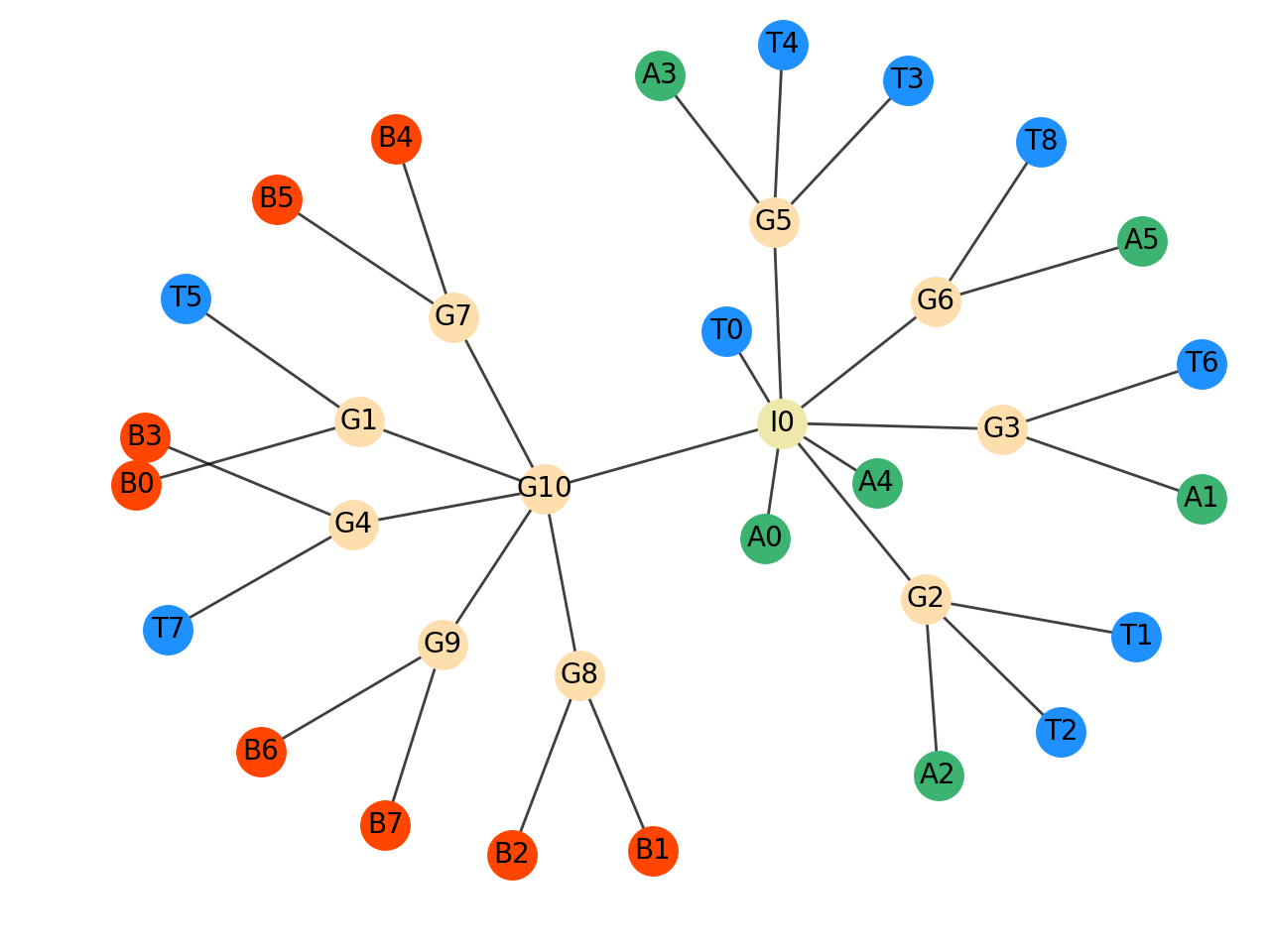} 
\includegraphics[width=0.5\textwidth]{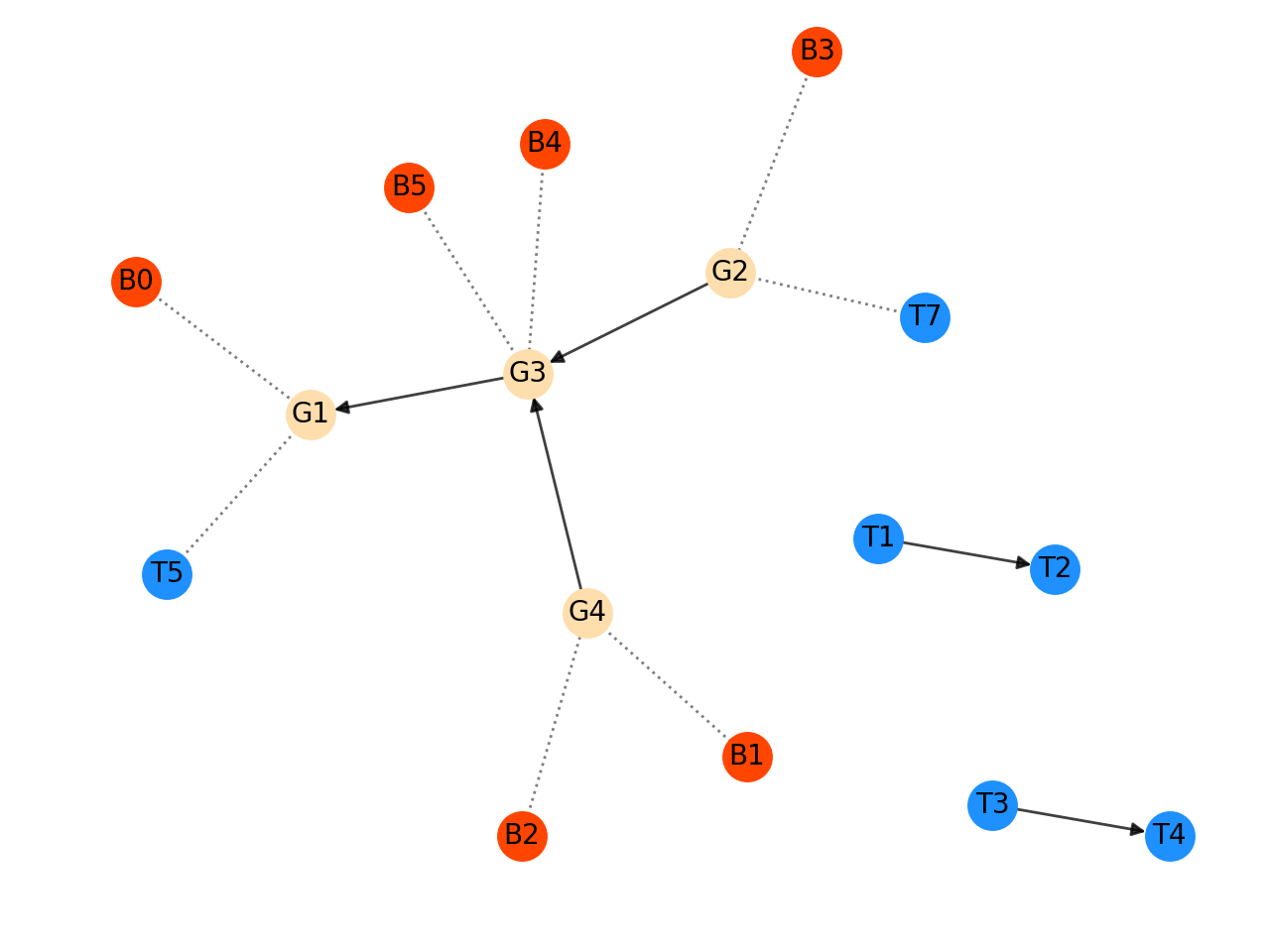} 
\includegraphics[width=0.5\textwidth]{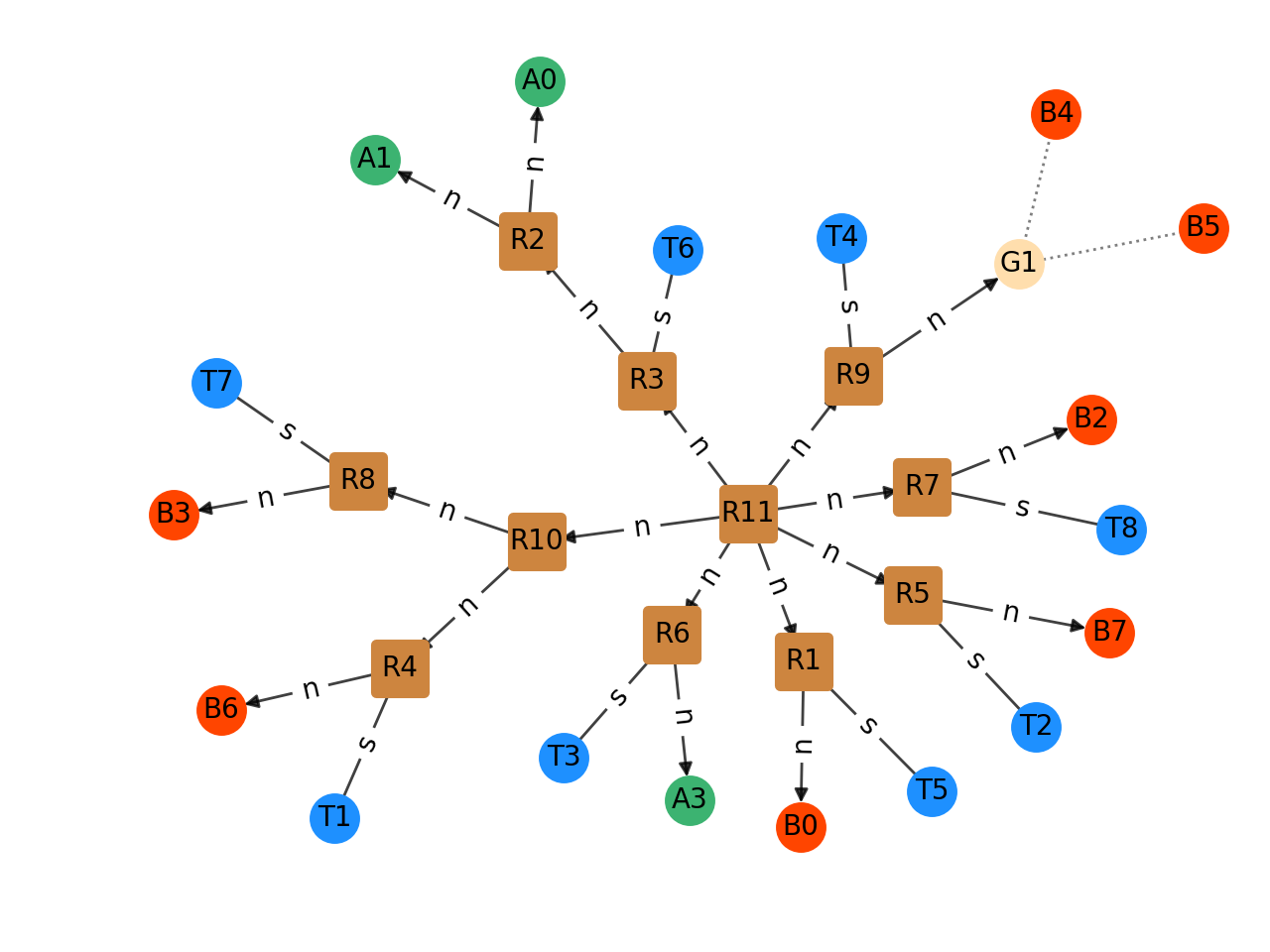} 
\caption{A discourse-driven decomposition of diagram layout (top left) with grouping (top right), connectivity (bottom left) and discourse structure (bottom right) graphs for diagram \#4210.}
\label{fig:ai2d-rst_mod_example}
\end{figure}

To solve the analytical problems described above, we propose an alternative, discourse-driven layout segmentation that overcomes the limitations discussed above by incorporating the distinctions provided by our definition of a semiotic mode (see Figure \ref{fig:semiotic_mode}). Figure \ref{fig:ai2d-rst_mod_example} shows a decomposition motivated by discourse structure for diagram \#4210, which picks out relevant parts of the cross-section. In contrast to the crowdsourced segmentation in Figure \ref{fig:ai2d_example}, the cross-section has been decomposed with the goal of maximising the coherence of discourse structure, which involves making available all the elements needed for such a representation of the diagram and its communicative intentions using the AI2D-RST annotation schema.

This is shown in Figure \ref{fig:ai2d-rst_mod_example}, which applies the AI2D-RST annotation schema to the diagram elements identified through discourse-driven decomposition. When provided with a sufficient inventory of diagram elements, the grouping graph more accurately reflects key structural properties of the diagram. The grouping graph (top right) contains two subgraphs, whose root nodes G10 and I0 correspond to the cross-section and cycle, respectively. Keeping in mind that the grouping graph seeks to capture visual groupings, this already provides a strong cue for two visually distinct configurations, which the AI2D-RST annotation schema refers to as \emph{macro-groups}. These constitute established configurations of the diagrammatic mode that may be flexibly combined in diagrams \cite[\p{681}]{hiippalaetal2019-ai2d}. To summarise, the grouping graph then already pulls these macro-groups apart and provides a foundation for their further analysis. We will shortly show how these macro-groups are integrated in the discourse structure graph.

The connectivity graph (bottom left) reveals that the diagram makes perhaps surprisingly limited use of arrows and lines as an expressive resource despite the intention that the diagram represents a cycle. This is one of the typical complicating factors contributing to the problems for annotation mentioned in the introduction above. The diagram does use arrows to set up connections between some individual elements and their groups, but the connectivity graph does not exhibit a cyclic structure. Some arrows, such as A2, have clear sources (T1; `Magma flows to surface ...') and targets (T2; `Weathering and erosion'), whereas other arrows, such as A4, do not. This encourages two alternative frames of interpretation for arrows \cite{alikhanistone2018}: some clearly signal transitions between stages (A2, A3), whereas others indicate the overall direction of the cycle (A4, A0).

The disconnections in the connectivity graph raise a crucial question: how does an interpretation involving a cyclic structure emerge if it is not clearly signalled using arrows? The answer to this question lies in the discourse structure of the graph as a whole, which here relies largely on \emph{written} language as an expressive resource. This allows the diagram to describe stages of the rock cycle explicitly using clausal structures, e.g. ``Metamorphic rock forms from heat and pressure'', but does not express the relationships diagrammatically using arrows. The verbal descriptions are instead placed in relation with specific regions of the cross-section, as shown in the discourse structure graph (bottom right).

The discourse structure graph illustrates how the cross-section and the cycle, which form separate subgraphs in the grouping graph, are tightly integrated in the discourse structure graph, capturing their \emph{joint} contribution towards a shared communicative goal and moving beyond the information in visual grouping alone. The specific rhetorical relations in Figure \ref{fig:ai2d-rst_mod_example} and criteria for their application, based loosely on Bateman \cite[\pp{149--162}]{bateman2008}, are given in abbreviated form in Table \ref{tab:rels}. Beginning from the top of the table, several \textsc{identification} relations are used to name regions (R1) and arrows (R6, R3). In relation R3, identification is extended to both arrows A0 and A1, which are joined together using the \textsc{joint} relation R2. \textsc{elaboration} relations R4--R5 and R7--R9, which assign descriptions to specific regions of the cross-section. This explains most of the phenomena depicted in the diagram. 
\begin{table}
\caption{Rhetorical relations in the discourse structure graph in Figure \ref{fig:ai2d-rst_mod_example}}
\begin{tabular}{p{2cm} p{2.6cm} p{3.8cm} p{3.3cm}}
\hline\noalign{\smallskip}
Identifier(s) & Relation & Nucleus & Satellite \\
\hline\noalign{\smallskip}
R1, R3, R6 & \textsc{identification} & Identified & Identifier \\
R2 & \textsc{joint} & No constraints & No constraints \\
R4--5, R7--9 & \textsc{elaboration} & Basic information & Additional information \\
R10 & \textsc{disjunction} & \raggedright Two or more alternatives & -- \\
R11 & \textsc{cyclic sequence} & Repeated steps & -- \\
\hline
\end{tabular}
\label{tab:rels}
\end{table}

All of these descriptions contribute towards an interpretation involving a cycle, which requires not only world knowledge, but is also supported using cohesive ties between lexical elements, such as the nouns `magma' and `rock' and the verb `to form'. The cycle itself is represented by the \textsc{cyclic sequence} relation R11, which joins together the individual descriptions that form its steps. The cycle also includes two possible alternatives, that is, whether magma cools below or above ground to form rocks, which is also explicitly captured by the \textsc{disjunction} relation R10 visible in the figure.

This analysis illustrates several of the methodological benefits of adopting a discourse-driven approach to unpacking the structure of diagrammatic representations. We can now move in a principled fashion beyond visual grouping and the individual sources of information in any diagram analysed to produce classifications more sensitive to the likely functions of the diagram as a whole. 

\section{Discussion}

We now briefly discuss some of the principal implications of our analysis for diagrams research more generally. The analysis has shown how  a multimodal perspective can yield valuable insights into diagrammatic representations by drawing on the broader basis provided by an appropriately differentiating view of the  diagrammatic semiotic mode. Instead of building pre-defined inventories of diagrammatic elements, for example, which are rapidly exhausted when faced with data that do not fall neatly into the categories defined, one can focus more on mapping the expressive resources available to the diagrammatic semiotic mode and describing the kinds of discourse structures they participate in.

This can be approached both empirically and with respect to existing proposals for the graphical elements and properties of diagrams. A recent example of such a proposal is that of Engelhardt and Richards, who seek to define ``universal building blocks of all types of diagrams and information graphics'' \cite[\p{201}]{EngelhardtRichards2018}. However, this still excludes ``context-related aspects'' of diagram use \cite[\p{203}]{EngelhardtRichards2018}, which, as we have seen above,  can be problematic when characterising larger collections of diagrams. A multimodal perspective is inherently geared towards addressing all of the aforementioned aspects of diagrammatic representations and naturally spans from form to contextually-motivated use. Furthermore, such frameworks can be applied reliably to diagrams, as exemplified by substantial inter-annotator agreement achieved for the AI2D-RST corpus \cite[\pp{674--679}]{hiippalaetal2019-ai2d}.

Multimodality research can also contribute towards a deeper understanding of \emph{signification} in diagrams, as this is precisely the work that expressive resources perform as part of the diagrammatic mode. As our analysis shows, diagrams that represent cycles do not necessarily need to draw on arrows for this purpose: the diagrammatic mode provides alternatives, such as written language, whose structural features (here: cohesive ties) may be used to cue a discourse semantic interpretation involving cyclicity. This allows a fine-grained decomposition of the proposed building blocks of diagrammatic representations \cite{hullmanbach2018,EngelhardtRichards2018}. Conversely, multimodality research is likely to benefit from the concepts developed in diagrams research for producing systematic descriptions of expressive resources. This will, however, require a significant effort in triangulating what has been done previously in multimodality and diagrams research, and aligning their theoretical concepts as necessary. Previous approaches to diagram classification as described above are the logical place to start such investigations.

Finally, our findings also carry implications for the computational modelling of diagrams. In particular, problems with the AI2D annotation \cite{kembhavietal2016} echo the need remarked on for mathematical diagrams above by Johansen and colleagues for domain expertise in describing the diagrammatic mode in order to achieve a description that respects its specific features. When applied to diagrams, computer vision tasks such as instance-level semantic segmentation and visual question answering must acknowledge particular characteristics of the diagrammatic mode. They should not be based simply on assumptions concerning how such tasks are defined for processing pictorial representations, since pictures constitute a quite different family of semiotic modes and exhibit rather different properties. Particularly important here is the issue of the appropriate level of semantic segmentation, that is, to what extent the mode in question needs to be decomposed into its components. Developing appropriate descriptions of the diagrammatic mode for computational modelling is therefore a task that needs to involve research communities working on both diagrams and multimodality.

\section{Conclusion}

We have introduced a multimodal perspective on diagrammatic representations, and presented a description of the diagrammatic semiotic mode, exemplifying the proposed approach using two recent multimodal diagram corpora. Multimodal analysis involves decomposing diagrammatic representations into their component parts, and we have argued for supporting decompositions driven by discourse structure -- that is, what the diagrammatic representations attempt to communicate and how their organisations explicitly guide readers to candidate interpretations. Capturing segmentations of this kind explicitly in appropriately designed corpora ensures that the necessary diagrammatic elements are available for further analysis. We suggest that given the widespread use of diagrams and their variation in different domains, an extensive programme of corpus-driven research of the kind we have proposed is now essential for developing an empirically-motivated account of the diagrammatic semiotic mode.

\bibliographystyle{splncs04}
\bibliography{bibliography,00}

\end{document}